\pdfoutput=1
\documentclass[11pt]{article}

\usepackage[final]{acl}

\usepackage{times}
\usepackage{latexsym}
\usepackage[disable]{todonotes}

\usepackage{amssymb}
\usepackage{pifont}
\usepackage{listings}

\usepackage[T1]{fontenc}
\usepackage[utf8]{inputenc}

\usepackage{microtype}

\usepackage{inconsolata}

\title{Are LLMs classical or nonmonotonic reasoners? Lessons from generics}

\author{Alina Leidinger \\
    ILLC\\
  University of Amsterdam \\
  \texttt{a.j.leidinger@uva.nl} \\\And
  Robert van Rooij \\
  ILLC\\
  University of Amsterdam \\
  \texttt{r.a.m.vanrooij@uva.nl} \\\And
  Ekaterina Shutova\\
  ILLC\\
  University of Amsterdam \\
  \texttt{e.shutova@uva.nl}}

\begin{document}
\maketitle
% \blfootnote{$^\star$ Correspondence to: \href{a.j.leidinger@uva.nl}{a.j.leidinger@uva.nl}.}

\begin{abstract}
Recent scholarship on reasoning in LLMs has supplied evidence of impressive performance and flexible adaptation to machine generated or human feedback. Nonmonotonic reasoning, crucial to human cognition for navigating the real world, remains a challenging, yet understudied task. In this work, we study nonmonotonic reasoning capabilities of seven state-of-the-art LLMs in one abstract and one commonsense reasoning task featuring generics, such as `Birds fly', and exceptions, `Penguins don't fly'~(see Fig. \ref{fig:teaser}). 
While LLMs exhibit reasoning patterns in accordance with human nonmonotonic reasoning abilities, they fail to maintain stable beliefs on truth conditions of generics at the addition of supporting examples (`Owls fly') or unrelated information (`Lions have manes').
Our findings highlight pitfalls in attributing human reasoning behaviours to LLMs, as well as assessing general capabilities, while consistent reasoning remains elusive.\footnote{Resources available at: \url{https://github.com/aleidinger/nonmonotonic_reasoning_generics}} 

\end{abstract}

\section{Introduction}

\begin{figure}
    \centering
    \includegraphics[width=.4\textwidth]{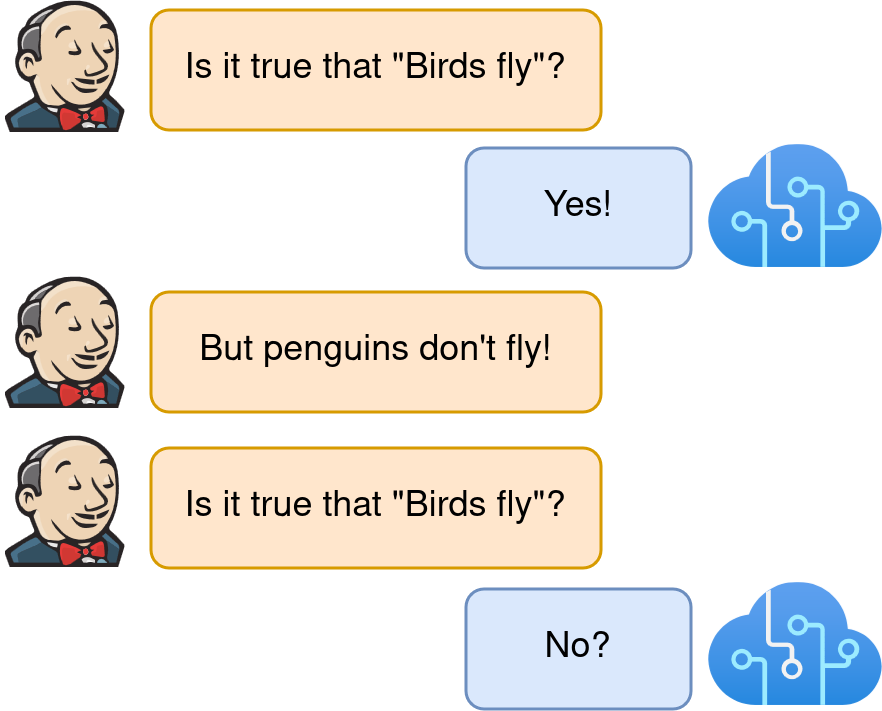}
    \caption{Reasoning about generics and exceptions}
    \label{fig:teaser}
\end{figure}

Generics are unquantified statements such as `Birds fly' or `Tigers are striped' \citep{carlson1995generic,mari2013genericity}. They are generalisations about kinds even if exceptions are known (`Penguins don't fly'; Fig.~\ref{fig:teaser}).  
Humans typically accept generics even if the property in question is rare among the kind~\citep[`Ticks carry the lime disease';][]{brandone2012lions,cimpian2010generic}. Generics play a crucial role in human beliefs on whether an example of a kind has a given property \citep{pelletier1997generics}. Human children master generics before they are able to reason about quantified statements \citep{hollander2002children,leslie2012quantified}.

In \textit{defeasible} or \textit{nonmonotonic} reasoning \citep{sloman2005problem,ginsberg1987readings,koons2005defeasible}, a hypothesis follows defeasibly from a premise, if the hypothesis is true in most \textit{normal} cases in which the premise holds. 
Generics make for a rich test bed for testing nonmonotonic reasoning capabilities~\citep{pelletier1997generics,asher1995some}. For example, given the generic `Birds fly' the inference `Tweety, the bird, can fly' is \textit{defeasibly valid}~\cite[i.a.]{mccarthy1986applications,reiter1988nonmonotonic}, i.e., it is reasonable to assume `Tweety can fly' even if exceptions are possible (`Tweety is a penguin')~\citep{lascarides1991discourse}. A classical reasoner however would reject the generic `Birds fly' upon learning that `Penguins don't fly'. 

Nonmonotonic reasoning is an integral part of human cognition \citep{russell2001cognitive}, that helps us to navigate the real-world, e.g., by \textit{planning}~\citep[Ch.5]{stenning2012human}, a task that LLMs still struggle with \citep{valmeekam2023planbench,stechly2024self}. 
Nonmonotonic reasoning poses a greater challenge for LLMs than other reasoning tasks \citep{han2024inductive} and hasn't been featured prominently among natural language inference (NLI) \citep{gubelmann2023capturing} or reasoning benchmarks (see \S \ref{relatedwork}).

The question of whether LLMs reason nonmonotonically or classically about generics and exceptions is intricately linked to desiderata of LLMs as reasoners.
LLMs are heralded for their ability to adapt to human or machine generated feedback~\citep[i.a.]{shinn2023reflexion,paul2023refiner,madaan2023self,pan2024automatically}. 
At the same time, it is desired that they reason \textit{reliably} when presented with invalid counterarguments, irrelevant information or user viewpoints. 
\textit{Sycophancy} \citep{perez-etal-2023-discovering} of LLMs, i.e., susceptibility to be swayed by user belief, is a case in point that has been investigated in recent studies~\cite[i.a.]{ranaldi2023sycophancy,laban2023you}.

As studies on reasoning patterns with generics remain scarce~\citep{ralethe2022generic,lin2020birds} and do not examine nonmonotonic reasoning, we address this gap by investigating the following \textit{research questions}: 1) Do LLMs reason nonmonotonically or classically about generics? 2) Are LLMs sensitive to counter-evidence in the form of exceptions? 3) Do LLMs reason consistently and reliably by maintaining their response given supporting or unrelated examples? 
We test seven state-of-the-art LLMs for their reasoning capabilities about generics in the presence of exceptions~(`Penguins don't fly'), as well as supporting~(`Owls fly') and irrelevant exemplars~(`Lions have manes'). Across two datasets featuring both abstract and commonsense generics, we find that LLM behaviour mirrors human nonmonotonic reasoning patterns in the presence of exceptions~(\S\ref{results_nonmonotonic}). However, most LLMs are not able to consistently maintain their agreement with generics given unrelated, or even supportive exemplars~(\S\ref{results_inconsistent}). Our study highlights challenges in comparing LLM behaviour to human reasoning patterns as well as assessing reasoning capabilities more broadly, while consistent reasoning cannot be guaranteed. In Section~\ref{discussion}, we present recommendations for a more holistic evaluation practice encompassing logical consistency measures. 

\section{Related Work}\label{relatedwork}

\subsection{Generics in NLP}

To date most works on generics focus on injecting commonsense knowledge or generics into LLMs~\cite[i.a.]{gajbhiye2022modelling,liu2023vera}, or training LLMs for knowledge/generic generation~\citep{bhagavatula-etal-2023-i2d2}. (See \citet{alkhamissi2022review} for a review.)
\citet{bhakthavatsalam2020genericskb} construct GenericsKG, a large knowledge base of generics as an asset for downstream tasks such as Question Answering or explanation generation. \citet{bhagavatula-etal-2023-i2d2} design a pipeline for synthetic generation of generics using samples from GenericsKB as seeds. \citet{allaway-etal-2023-penguins} in turn complement the data with exceptions and instantiations for each generic, but do not investigate nonmonotonic reasoning capabilities. 

Most closely related to our work, \citet{lin2020birds} find that LMs struggle to predict numerical knowledge in generics such as `Birds have two legs'. \citet{ralethe2022generic} find that pre-trained masked LMs falsely \textit{overgeneralise} \citep{leslie2011overgeneralisation} from generics (`Ducks lay eggs') to universally quantified statements (`All ducks lay eggs').

\subsection{Nonmonotonic reasoning in NLP}

\citet{han2024inductive} test nonmonotonic reasoning among other inductive reasoning tasks and find that only GPT-4 performs adequately. LLMs struggle to reason with contradictory information \citep{kazemi2023boardgameqa}. \citet{rudinger2020thinking,brahman2021learning,bhagavatula2019abductive} develop NLI tasks to test defeasible or abductive reasoning in pragmatics, while \citet{pyatkin-etal-2023-clarifydelphi,ziems-etal-2023-normbank,rao2023makes} focus on defeasible reasoning and social norms. \citet{parmar2024towards} introduce non-monotonic reasoning tasks inspired by \citet{lifschitz1989benchmark} as part of their LogicBench.

\subsection{Consistency in reasoning}

Most recent studies on reliability and consistency in reasoning examine sycophancy \citep{perez-etal-2023-discovering,laban2023you,ranaldi2023sycophancy}, consistency within multi-step reasoning or across sessions and users \citep{chen2023two,wang2022self}. (See \citet{liu2023trustworthy} for a review.)

Orthogonal to this, our work connects to studies of reasoning in the presence of unrelated or conflicting information.
\citet{shi2023large} find that LLMs are easily confounded by irrelevant information in arithmetic reasoning. 
Across a variety of reasoning tasks, \citet{wang2023can} find that OpenAI models struggle to maintain stable responses given irrelevant objections.
\citet{xie2023adaptive} find mixed evidence of LLMs being sensitive to information that contradicts prior knowledge, yet showing a form of `confirmation bias' when presented with diverse viewpoints.

\section{Tasks and datasets}

We test nonmonotonic reasoning with generics using two datasets, featuring commonsense and abstract generics. Both datasets contain generics~(`Birds fly') accompanied by statements where the generic holds~(`Owls fly') or doesn't~(`Penguins don't fly'). We refer to such examples as \textit{instantiations} or \textit{exceptions} respectively, and to both collectively as as \textit{exemplars}.

As commonsense generics, we use the synthetic dataset of generics and exemplars released by~\citet{allaway-etal-2023-penguins} (henceforth referred to as \texttt{GEN-comm}). %\footnote{The data is publicly available at \url{https://github.com/emilyallaway/generics-exemplars}.} 
The dataset consists of $\sim650$ generics and $\sim19.000$ exemplars (E.g., `Hoes are used to plow fields or clear snow'; `Hoes can be used to cut grass').\footnote{See App. \ref{sec:preprocessing} for additional information on preprocessing.}
Secondly, we construct an abstract reasoning dataset featuring generics (\texttt{GEN-abs}). Inspired by \citet{han2024inductive}, we use categories (`birds') and examples (`eagles') from \citet{dedeyne2008exemplar} to construct generics of the form `Birds have property P' and exemplars of the form `Eagles do (not) have property P'. The dataset contains 260 tuples of a generic paired with an exemplar.\footnote{The dataset is available at: \url{https://github.com/aleidinger/nonmonotonic_reasoning_generics/blob/main/data/abstract_generics.csv}}

For both datasets, our goal is to prompt LLMs for their agreement with a generic in the presence of exemplars which confirm or contradict the generic. We use the following prompt template, including model-specific special tokens\footnote{See Appendix \ref{sec:example_input} or \url{https://huggingface.co/docs/transformers/main/en/chat_templating} for details.} to signal a chat history between an assistant and a user.\footnote{We also experiment with an alternative prompting template and Chain-of-Thought prompting. Since results are similar, they are included in Appendix \ref{appendix_additional_results}.}

\noindent\fbox{\begin{minipage}{19em}
\underline{Example:}

\textit{[INST] Is the following statement true: ``Birds fly.'' \textbackslash nPlease answer yes or no. [/INST]}\\
\textit{yes}\\
\textit{[INST] Penguins don't fly.\textbackslash nIs the following statement true: ``Birds fly.''\textbackslash nPlease answer yes or no. [/INST]}
\end{minipage}}\\

As a control study, we also replace the exception in the prompt (\textit{`Penguins don't fly'}) with an instantiation (\textit{`Owls fly'}) or a random exemplar~(\textit{`Hoes can be used to cut grass'}).
Since generics in \texttt{GEN-abs} are abstract in nature, and to enable a consistent set-up across both datasets, we retain generics in \texttt{GEN-comm} that LLMs accepts when prompted with the first part of the above template, e.g., 
\textit{[INST] Is the following statement true: ``Birds fly.'' \textbackslash nPlease answer yes or no. [/INST]}.\footnote{See App. \ref{sec:preprocessing} for details and results on discarded generics.}

\section{Method}\label{method}

\subsection{Models}\label{models}

We conduct our experiments on medium-sized open-weight models selected from the top of AlpacaEval\footnote{\url{https://tatsu-lab.github.io/alpaca_eval/}} and LMSys\footnote{\url{https://chat.lmsys.org/?leaderboard}} leaderboards, namely Llama-2-13b \citep{touvron2023llama}, Mistral-7b-Instruct-v0.2 \citep{jiang2023mistral}, Mixtral-8x7B-Instruct-v0.1 \citep{jiang2024mixtral}, Zephyr-7b-beta \citep{tunstall2023zephyr}, WizardLM-13B-V1.2 \citep{xu2023wizardlm}, Starling-LM-7B-alpha~\citep{starling2023}, and OpenHermes-2.5-Mistral-7B \citep{openhermes}.\footnote{See App. \ref{sec:appendix_models} for checkpoints and additional information.} 

\subsection{Prompting set-up}

Since LLM behaviour can vary considerably with the phrasing of an instruction \citep{webson-pavlick-2022-prompt, leidinger2023language}, we formulate three different instructions to test if an LLM agrees with a given generic: \textit{`Is the following statement true', `Do you believe the following statement to be true', `Do you believe that the following statement is accurate'}. 
Since the optimal model reply is short and succinct, we follow the convention of HELM~\cite[p.161]{liang2022helm} in setting temperature to $0$ for reproducibility across runs. 
We format every prompt using the chat template appropriate for each model, with no system prompt.\footnotemark[4]
To map LLM responses to labels \texttt{disagree} vs. \texttt{agree}, we use pattern matching and record whether a response starts with \textit{yes} or \textit{no} \citep{rottger2023xstest}.
We aggregate responses for the three instructions via majority voting.

\subsection{Statistical tests}
To assess whether behaviour of LLMs is significantly different in the absence vs. presence of exemplars we resort to non-parametric statistical testing. Since our samples are paired, we use the Wilcoxon signed-rank test \citep{wilcoxon1992individual}. % and the Friedman test \citep{friedman1937use}.

\begin{figure}[!t]
    \centering
    \begin{minipage}[b]{0.47\textwidth}
    \includegraphics[width=\textwidth]{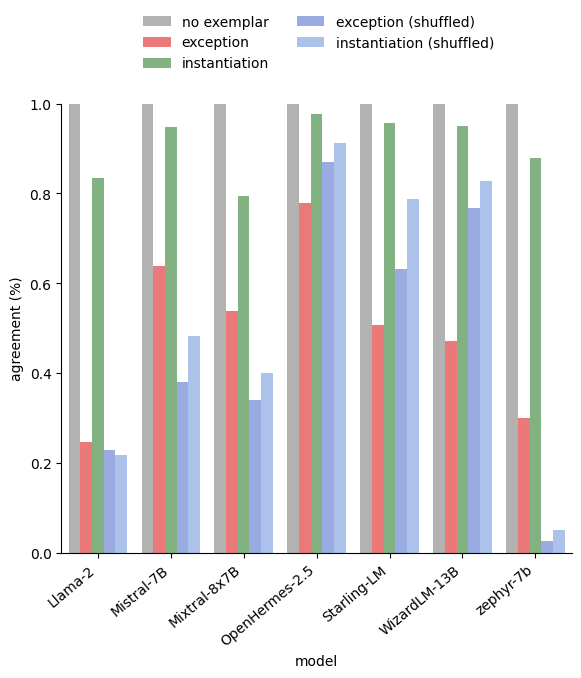}
    \end{minipage}

    \begin{minipage}[b]{0.47\textwidth}
    \includegraphics[width=\textwidth]{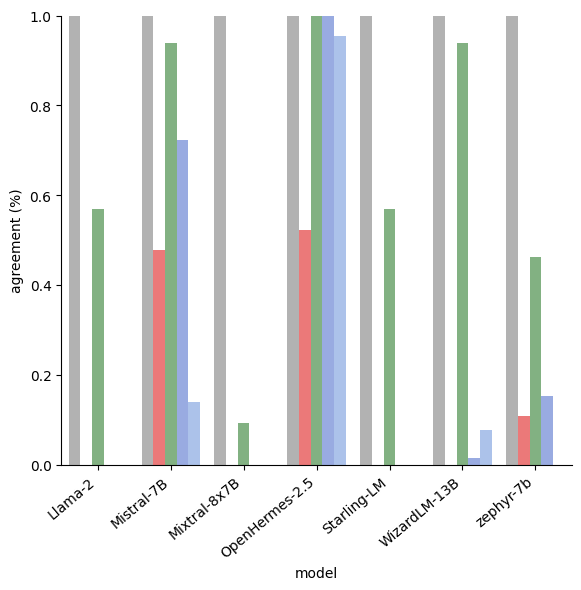}
    \end{minipage}
    \caption{LLM agreement with generics in the presence of exemplars on \texttt{GEN-comm} (top) and \texttt{GEN-abs} (bottom). Missing columns indicate agreement rates of $0\%$.}
        \label{fig:results}
\end{figure}

\section{Results}\label{results}

We present our main results in Figure \ref{fig:results}. Additional, accordant results are described in Appendix \ref{appendix_additional_results}.

\subsection{Do LLMs reason nonmonotonically?}\label{results_nonmonotonic}

Since humans maintain their beliefs about truth conditions of generics (`Birds fly') in the presence of exceptions (`Penguins do not fly'), we examine whether challenging LLMs with an exception decreases their agreement to generics significantly. We find this to be the case for all models on both datasets ($p=0.01$; see App. \ref{app:statisticaltestresults} for statistical test results). Notably, agreement rates drop to 0 for Llama-2, Mixtral, Starling and WizardLM on \texttt{GEN-abs}.

\subsection{Do LLMs reason consistently?}\label{results_inconsistent}
In the presence of supporting evidence (\textit{instantiation}) to a generic (`Owls fly'), we expect LLM agreement to remain at $100\%$, but this is not the case. While agreement rates remain high in numbers, they drop significantly for all models. On \texttt{GEN-abs}, only Mistral, OpenHermes, and WizardLM maintain agreement rates of $>90\%$, while agreement drops to $<10\%$ for Mixtral.

Similarly, most LLMs are not able to disregard irrelevant random exemplars (\textit{exception/instantiation (shuffled))}. Agreement rates decline steeply below $50\%$ for Llama-2, Mistral, Mixtral and Zephyr on \texttt{GEN-comm} and to below $20\%$ for Llama-2, Mixtral, Starling, WizardLM and Zephyr on \texttt{GEN-abs}. OpenHermes stands out as the only model that maintains agreement rates above $85\%$ on both datasets. Notably, OpenHermes is the only model which has been trained on additional code data which has been shown to also help reasoning in natural language \citep{liang2022helm,yang2024if,ma2023training}. Nevertheless, observed differences are statistically significant for all models on both datasets (App. \ref{app:statisticaltestresults}).

\begin{figure*}
    \centering
    \includegraphics[width=.7\textwidth]{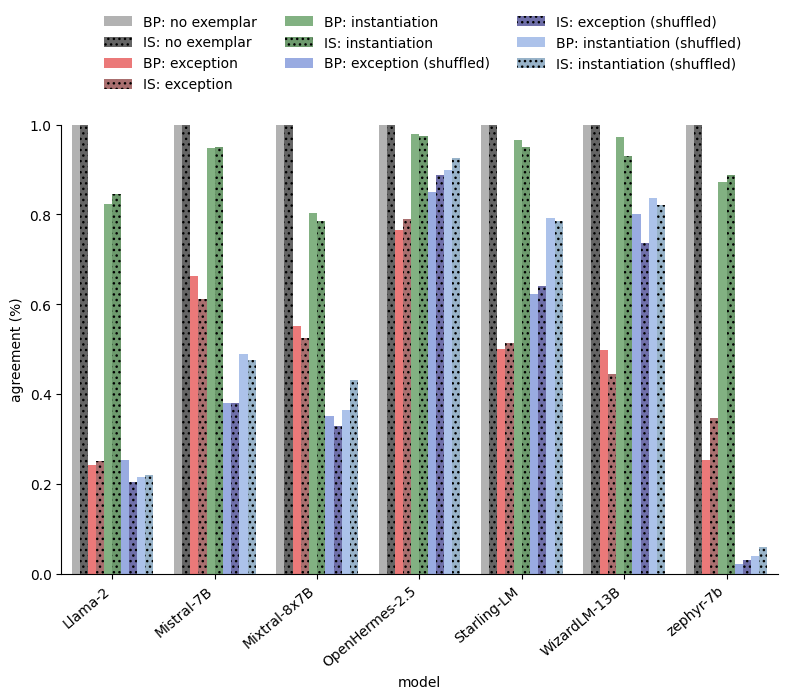}
    \caption{LLM agreement with bare plural (BP) and indefinite singular (IS) generics in the presence of exemplars on \texttt{GEN-comm}.}
    \label{fig:generic-kinds}
\end{figure*}

\section{Analysis}

\subsection{How do LLMs reason about different types of generics?}

\texttt{GEN-comm} contains both bare plural (BP) generics as well as indefinite singular (IS) generics \citep{leslie2009conceptual}. (For example, `Sea snails have a hard shell, which protects them from predators' (BP) and `A deciduous tree can be identified by its leaves' (IS)). We did not find notable differences between LLM agreement to BP or IS generics in the presence of exemplars (see Figure \ref{fig:generic-kinds}). Aforementioned consistency failures persist for both types of generics. 

\subsection{Qualitative analysis}

Generics in \texttt{GEN-comm} which are accepted in isolation, but are rejected in the presence of exceptions or instantiations include `Stimulants can be used to treat ADHD' (Llama-2, Starling, Mixtral) or `A bobsleigh is driven by a single driver' (Starling, Mistral, Mixtral, OpenHermes, WizardLM). Generics which are accepted no matter the exemplar presented in context include `Inflammatory diseases may be caused by an imbalance of the immune system' (Llama-2, Starling, Mistral, OpenHermes), `A processor should be able to run a program' (Starling, Mixtral, OpenHermes, WizardLM), `Experimental evidence is used to support or refute theories', `An adventure has a beginning, middle, end' (Starling, OpenHermes, WizardLM), and `Coincidence is part of the human condition' (Starling, Mistral, OpenHermes).

For \texttt{GEN-abs}, OpenHermes is the only LLM which maintains its agreement to a generic (`Birds have property P', `Mammals have property P') in the presence of any instantiation or unrelated exemplar, but flips its decision and outputs disagreement in the presence of an exception. No LLM accepts any of the generics regardless of the exemplar it is paired with. 

\section{Discussion}\label{discussion}

With the advent of LLMs and reports of impressive performance, including on reasoning tasks \citep{wei2022finetuned,kojima2022large}, recent investigations into failure modes in reasoning have focused, e.g., on prompt attacks \citep[i.a.]{zhu2023promptbench,wang2023robustness}, sycophancy~\citep[i.a.]{perez-etal-2023-discovering,laban2023you,ranaldi2023sycophancy} or adaptability to critique or feedback \citep{madaan2023self,chen2023reconcile,huang2023large,pan2024automatically}. Such research trends might be seen as emblematic of a view of LLMs as artificial natural artifacts \citep{subbarao2022}. Results in this study demonstrate the difficulties of making claims about reasoning capabilities of LLMs or comparing them to human reasoners \citep{han2024inductive,ralethe2022generic,lin2020birds}, while consistent reasoning remains elusive even for state-of-the-art LLMs. 
Research that predates the paradigm shift to few-shot prompting, has advocated for arguably simpler, systematic diagnostic tests \citep{ribeiro-etal-2020-beyond,ettinger-2020-bert,kassner2020negated}. We argue that such behavioural tests merit a revival, so that performance metrics for reasoning are complemented with measures of logical consistency and robustness. %We would like to advocate 

\section{Conclusion}

The present study focuses on nonmonotonic reasoning capabilities of LLMs in the context of generics. We evaluate seven state-of-the-art LLMs on two datasets featuring both abstract and commonsense generic statements. While LLM behaviour on generics paired with exceptions is in line with nonmonotonic reasoning patterns, LLMs fail to reason consistently and robustly when adding supporting or unrelated exemplars. 

\section{Limitations}

We acknowledge that our experiments exclusively feature generics and exemplars in English. Future research might profit from including additional languages to examine nonmonotonic reasoning capabilities in other languages, drawing on cross-linguistic research on generics \citep{mari2013genericity}. Such work might also highlight differences in consistency failures between different languages. 
In this work, we do not experiment with generics pertaining to demographic groups or nationalities because of concerns around social bias.
Future work might examine LLM behaviour on generic statements for larger LLMs or closed-source models. We restrict ourselves to medium-sized open-weight LLMs, due to their widespread use and availability, as well as restrictions on our computational budget.

\section*{Acknowledgements}
We thank our anonymous reviewers for their insightful comments. The work for this publication is financially supported by the project, `From Learning to Meaning: A new approach to Generic Sentences and Implicit Biases' (project number 406.18.TW.007) of the research programme SGW Open Competition, which is (partly) financed by the Dutch Research Council (NWO).

\bibliography{anthology,custom}

\begin{thebibliography}{78}
\expandafter\ifx\csname natexlab\endcsname\relax\def\natexlab#1{#1}\fi

\bibitem[{AlKhamissi et~al.(2022)AlKhamissi, Li, Celikyilmaz, Diab, and Ghazvininejad}]{alkhamissi2022review}
Badr AlKhamissi, Millicent Li, Asli Celikyilmaz, Mona Diab, and Marjan Ghazvininejad. 2022.
\newblock A review on language models as knowledge bases.
\newblock \emph{arXiv preprint arXiv:2204.06031}.

\bibitem[{Allaway et~al.(2023)Allaway, Hwang, Bhagavatula, McKeown, Downey, and Choi}]{allaway-etal-2023-penguins}
Emily Allaway, Jena~D. Hwang, Chandra Bhagavatula, Kathleen McKeown, Doug Downey, and Yejin Choi. 2023.
\newblock \href {https://doi.org/10.18653/v1/2023.eacl-main.192} {{P}enguins don{'}t fly: Reasoning about generics through instantiations and exceptions}.
\newblock In \emph{Proceedings of the 17th Conference of the European Chapter of the Association for Computational Linguistics}, pages 2618--2635, Dubrovnik, Croatia. Association for Computational Linguistics.

\bibitem[{Asher and Morreau(1995)}]{asher1995some}
Nicholas Asher and Michael Morreau. 1995.
\newblock What some generic sentences mean.
\newblock \emph{The generic book}, pages 300--338.

\bibitem[{Bhagavatula et~al.(2023)Bhagavatula, Hwang, Downey, Le~Bras, Lu, Qin, Sakaguchi, Swayamdipta, West, and Choi}]{bhagavatula-etal-2023-i2d2}
Chandra Bhagavatula, Jena~D. Hwang, Doug Downey, Ronan Le~Bras, Ximing Lu, Lianhui Qin, Keisuke Sakaguchi, Swabha Swayamdipta, Peter West, and Yejin Choi. 2023.
\newblock \href {https://doi.org/10.18653/v1/2023.acl-long.535} {{I}2{D}2: Inductive knowledge distillation with {N}euro{L}ogic and self-imitation}.
\newblock In \emph{Proceedings of the 61st Annual Meeting of the Association for Computational Linguistics (Volume 1: Long Papers)}, pages 9614--9630, Toronto, Canada. Association for Computational Linguistics.

\bibitem[{Bhagavatula et~al.(2019)Bhagavatula, Le~Bras, Malaviya, Sakaguchi, Holtzman, Rashkin, Downey, Yih, and Choi}]{bhagavatula2019abductive}
Chandra Bhagavatula, Ronan Le~Bras, Chaitanya Malaviya, Keisuke Sakaguchi, Ari Holtzman, Hannah Rashkin, Doug Downey, Wen-tau Yih, and Yejin Choi. 2019.
\newblock Abductive commonsense reasoning.
\newblock In \emph{International Conference on Learning Representations}.

\bibitem[{Bhakthavatsalam et~al.(2020)Bhakthavatsalam, Anastasiades, and Clark}]{bhakthavatsalam2020genericskb}
Sumithra Bhakthavatsalam, Chloe Anastasiades, and Peter Clark. 2020.
\newblock Genericskb: A knowledge base of generic statements.
\newblock \emph{arXiv preprint arXiv:2005.00660}.

\bibitem[{Brahman et~al.(2021)Brahman, Shwartz, Rudinger, and Choi}]{brahman2021learning}
Faeze Brahman, Vered Shwartz, Rachel Rudinger, and Yejin Choi. 2021.
\newblock Learning to rationalize for nonmonotonic reasoning with distant supervision.
\newblock In \emph{Proceedings of the AAAI Conference on Artificial Intelligence}, volume~35, pages 12592--12601.

\bibitem[{Brandone et~al.(2012)Brandone, Cimpian, Leslie, and Gelman}]{brandone2012lions}
Amanda~C Brandone, Andrei Cimpian, Sarah-Jane Leslie, and Susan~A Gelman. 2012.
\newblock Do lions have manes? for children, generics are about kinds rather than quantities.
\newblock \emph{Child development}, 83(2):423--433.

\bibitem[{Carlson and Pelletier(1995)}]{carlson1995generic}
Gregory~N Carlson and Francis~Jeffry Pelletier. 1995.
\newblock \emph{The generic book}.
\newblock University of Chicago Press.

\bibitem[{Chen et~al.(2023{\natexlab{a}})Chen, Phang, Parrish, Padmakumar, Zhao, Bowman, and Cho}]{chen2023two}
Angelica Chen, Jason Phang, Alicia Parrish, Vishakh Padmakumar, Chen Zhao, Samuel~R Bowman, and Kyunghyun Cho. 2023{\natexlab{a}}.
\newblock Two failures of self-consistency in the multi-step reasoning of llms.
\newblock \emph{Transactions on Machine Learning Research}.

\bibitem[{Chen et~al.(2023{\natexlab{b}})Chen, Saha, and Bansal}]{chen2023reconcile}
Justin Chih-Yao Chen, Swarnadeep Saha, and Mohit Bansal. 2023{\natexlab{b}}.
\newblock Reconcile: Round-table conference improves reasoning via consensus among diverse llms.
\newblock \emph{arXiv preprint arXiv:2309.13007}.

\bibitem[{Cimpian et~al.(2010)Cimpian, Brandone, and Gelman}]{cimpian2010generic}
Andrei Cimpian, Amanda~C Brandone, and Susan~A Gelman. 2010.
\newblock Generic statements require little evidence for acceptance but have powerful implications.
\newblock \emph{Cognitive science}, 34(8):1452--1482.

\bibitem[{De~Deyne et~al.(2008)De~Deyne, Verheyen, Ameel, Vanpaemel, Dry, Voorspoels, and Storms}]{dedeyne2008exemplar}
Simon De~Deyne, Steven Verheyen, Eef Ameel, Wolf Vanpaemel, Matthew~J Dry, Wouter Voorspoels, and Gert Storms. 2008.
\newblock Exemplar by feature applicability matrices and other dutch normative data for semantic concepts.
\newblock \emph{Behavior research methods}, 40:1030--1048.

\bibitem[{Ettinger(2020)}]{ettinger-2020-bert}
Allyson Ettinger. 2020.
\newblock \href {https://doi.org/10.1162/tacl_a_00298} {What {BERT} is not: Lessons from a new suite of psycholinguistic diagnostics for language models}.
\newblock \emph{Transactions of the Association for Computational Linguistics}, 8:34--48.

\bibitem[{Gajbhiye et~al.(2022)Gajbhiye, Anke, and Schockaert}]{gajbhiye2022modelling}
Amit Gajbhiye, Luis~Espinosa Anke, and Steven Schockaert. 2022.
\newblock Modelling commonsense properties using pre-trained bi-encoders.
\newblock In \emph{Proceedings of the 29th International Conference on Computational Linguistics}, pages 3971--3983.

\bibitem[{Ginsberg(1987)}]{ginsberg1987readings}
Matthew~L Ginsberg. 1987.
\newblock Readings in nonmonotonic reasoning.

\bibitem[{Gubelmann et~al.(2023)Gubelmann, Katis, Niklaus, and Handschuh}]{gubelmann2023capturing}
Reto Gubelmann, Ioannis Katis, Christina Niklaus, and Siegfried Handschuh. 2023.
\newblock Capturing the varieties of natural language inference: A systematic survey of existing datasets and two novel benchmarks.
\newblock \emph{Journal of Logic, Language and Information}, pages 1--28.

\bibitem[{Han et~al.(2024)Han, Ransom, Perfors, and Kemp}]{han2024inductive}
Simon~Jerome Han, Keith~J Ransom, Andrew Perfors, and Charles Kemp. 2024.
\newblock Inductive reasoning in humans and large language models.
\newblock \emph{Cognitive Systems Research}, 83:101155.

\bibitem[{Hollander et~al.(2002)Hollander, Gelman, and Star}]{hollander2002children}
Michelle~A Hollander, Susan~A Gelman, and Jon Star. 2002.
\newblock Children's interpretation of generic noun phrases.
\newblock \emph{Developmental psychology}, 38(6):883.

\bibitem[{Huang et~al.(2023)Huang, Chen, Mishra, Zheng, Yu, Song, and Zhou}]{huang2023large}
Jie Huang, Xinyun Chen, Swaroop Mishra, Huaixiu~Steven Zheng, Adams~Wei Yu, Xinying Song, and Denny Zhou. 2023.
\newblock Large language models cannot self-correct reasoning yet.
\newblock In \emph{The Twelfth International Conference on Learning Representations}.

\bibitem[{Jiang et~al.(2023)Jiang, Sablayrolles, Mensch, Bamford, Chaplot, Casas, Bressand, Lengyel, Lample, Saulnier et~al.}]{jiang2023mistral}
Albert~Q Jiang, Alexandre Sablayrolles, Arthur Mensch, Chris Bamford, Devendra~Singh Chaplot, Diego de~las Casas, Florian Bressand, Gianna Lengyel, Guillaume Lample, Lucile Saulnier, et~al. 2023.
\newblock Mistral 7b.
\newblock \emph{arXiv preprint arXiv:2310.06825}.

\bibitem[{Jiang et~al.(2024)Jiang, Sablayrolles, Roux, Mensch, Savary, Bamford, Chaplot, Casas, Hanna, Bressand et~al.}]{jiang2024mixtral}
Albert~Q Jiang, Alexandre Sablayrolles, Antoine Roux, Arthur Mensch, Blanche Savary, Chris Bamford, Devendra~Singh Chaplot, Diego de~las Casas, Emma~Bou Hanna, Florian Bressand, et~al. 2024.
\newblock Mixtral of experts.
\newblock \emph{arXiv preprint arXiv:2401.04088}.

\bibitem[{Kambhampati(2022)}]{subbarao2022}
Subbarao Kambhampati. 2022.
\newblock \href {https://cacm.acm.org/blogs/blog-cacm/261732-ai-as-an-ersatz-natural-science/fulltext} {Ai as (an ersatz) natural science?}

\bibitem[{Kassner and Sch{\"u}tze(2020)}]{kassner2020negated}
Nora Kassner and Hinrich Sch{\"u}tze. 2020.
\newblock \href {https://doi.org/10.18653/v1/2020.acl-main.698} {Negated and misprimed probes for pretrained language models: Birds can talk, but cannot fly}.
\newblock In \emph{Proceedings of the 58th Annual Meeting of the Association for Computational Linguistics}, pages 7811--7818, Online. Association for Computational Linguistics.

\bibitem[{Kazemi et~al.(2024)Kazemi, Yuan, Bhatia, Kim, Xu, Imbrasaite, and Ramachandran}]{kazemi2023boardgameqa}
Mehran Kazemi, Quan Yuan, Deepti Bhatia, Najoung Kim, Xin Xu, Vaiva Imbrasaite, and Deepak Ramachandran. 2024.
\newblock Boardgameqa: A dataset for natural language reasoning with contradictory information.
\newblock \emph{Advances in Neural Information Processing Systems}, 36.

\bibitem[{Kojima et~al.(2022)Kojima, Gu, Reid, Matsuo, and Iwasawa}]{kojima2022large}
Takeshi Kojima, Shixiang~Shane Gu, Machel Reid, Yutaka Matsuo, and Yusuke Iwasawa. 2022.
\newblock Large language models are zero-shot reasoners.
\newblock \emph{Advances in neural information processing systems}, 35:22199--22213.

\bibitem[{Koons(2005)}]{koons2005defeasible}
Robert Koons. 2005.
\newblock Defeasible reasoning.

\bibitem[{Laban et~al.(2023)Laban, Murakhovs'~ka, Xiong, and Wu}]{laban2023you}
Philippe Laban, Lidiya Murakhovs'~ka, Caiming Xiong, and Chien-Sheng Wu. 2023.
\newblock Are you sure? challenging llms leads to performance drops in the flipflop experiment.
\newblock \emph{arXiv preprint arXiv:2311.08596}.

\bibitem[{Lascarides and Asher(1991)}]{lascarides1991discourse}
Alex Lascarides and Nicholas Asher. 1991.
\newblock Discourse relations and defeasible knowledge.
\newblock In \emph{29th Annual Meeting of the Association for Computational Linguistics}, pages 55--62.

\bibitem[{Leidinger et~al.(2023)Leidinger, van Rooij, and Shutova}]{leidinger2023language}
Alina Leidinger, Robert van Rooij, and Ekaterina Shutova. 2023.
\newblock \href {https://doi.org/10.18653/v1/2023.findings-emnlp.618} {The language of prompting: What linguistic properties make a prompt successful?}
\newblock In \emph{Findings of the Association for Computational Linguistics: EMNLP 2023}, pages 9210--9232, Singapore. Association for Computational Linguistics.

\bibitem[{Leslie and Gelman(2012)}]{leslie2012quantified}
Sarah-Jane Leslie and Susan~A Gelman. 2012.
\newblock Quantified statements are recalled as generics: Evidence from preschool children and adults.
\newblock \emph{Cognitive psychology}, 64(3):186--214.

\bibitem[{Leslie et~al.(2011)Leslie, Khemlani, and Glucksberg}]{leslie2011overgeneralisation}
Sarah-Jane Leslie, Sangeet Khemlani, and Sam Glucksberg. 2011.
\newblock \href {https://doi.org/https://doi.org/10.1016/j.jml.2010.12.005} {{Do all ducks lay eggs? The generic overgeneralization effect}}.
\newblock \emph{Journal of Memory and Language}, 65(1):15--31.

\bibitem[{Leslie et~al.(2009)Leslie, Khemlani, Prasada, and Glucksberg}]{leslie2009conceptual}
Sarah-Jane Leslie, Sangeet Khemlani, Sandeep Prasada, and Sam Glucksberg. 2009.
\newblock Conceptual and linguistic distinctions between singular and plural generics.
\newblock \emph{Proceedings of the 31st annual cognitive science society}, pages 479--484.

\bibitem[{Liang et~al.(2023)Liang, Bommasani, Lee, Tsipras, Soylu, Yasunaga, Zhang, Narayanan, Wu, Kumar et~al.}]{liang2022helm}
Percy Liang, Rishi Bommasani, Tony Lee, Dimitris Tsipras, Dilara Soylu, Michihiro Yasunaga, Yian Zhang, Deepak Narayanan, Yuhuai Wu, Ananya Kumar, et~al. 2023.
\newblock Holistic evaluation of language models.
\newblock \emph{Transactions on Machine Learning Research}.

\bibitem[{Lifschitz(1989)}]{lifschitz1989benchmark}
Vladimir Lifschitz. 1989.
\newblock Benchmark problems for formal nonmonotonic reasoning: Version 2.00.
\newblock In \emph{Non-Monotonic Reasoning: 2nd International Workshop Grassau, FRG, June 13--15, 1988 Proceedings 2}, pages 202--219. Springer.

\bibitem[{Lin et~al.(2020)Lin, Lee, Khanna, and Ren}]{lin2020birds}
Bill~Yuchen Lin, Seyeon Lee, Rahul Khanna, and Xiang Ren. 2020.
\newblock \href {https://doi.org/10.18653/v1/2020.emnlp-main.557} {{B}irds have four legs?! {N}umer{S}ense: {P}robing {N}umerical {C}ommonsense {K}nowledge of {P}re-{T}rained {L}anguage {M}odels}.
\newblock In \emph{Proceedings of the 2020 Conference on Empirical Methods in Natural Language Processing (EMNLP)}, pages 6862--6868, Online. Association for Computational Linguistics.

\bibitem[{Liu et~al.(2023{\natexlab{a}})Liu, Wang, Wang, Smith, Choi, and Hajishirzi}]{liu2023vera}
Jiacheng Liu, Wenya Wang, Dianzhuo Wang, Noah Smith, Yejin Choi, and Hannaneh Hajishirzi. 2023{\natexlab{a}}.
\newblock \href {https://doi.org/10.18653/v1/2023.emnlp-main.81} {Vera: A general-purpose plausibility estimation model for commonsense statements}.
\newblock In \emph{Proceedings of the 2023 Conference on Empirical Methods in Natural Language Processing}, pages 1264--1287, Singapore. Association for Computational Linguistics.

\bibitem[{Liu et~al.(2023{\natexlab{b}})Liu, Yao, Ton, Zhang, Guo, Cheng, Klochkov, Taufiq, and Li}]{liu2023trustworthy}
Yang Liu, Yuanshun Yao, Jean-Francois Ton, Xiaoying Zhang, Ruocheng Guo, Hao Cheng, Yegor Klochkov, Muhammad~Faaiz Taufiq, and Hang Li. 2023{\natexlab{b}}.
\newblock Trustworthy llms: a survey and guideline for evaluating large language models' alignment.
\newblock In \emph{Socially Responsible Language Modelling Research}.

\bibitem[{Ma et~al.(2023)Ma, Liu, Yu, Zhang, Jiang, Wang, and Li}]{ma2023training}
Yingwei Ma, Yue Liu, Yue Yu, Yuanliang Zhang, Yu~Jiang, Changjian Wang, and Shanshan Li. 2023.
\newblock At which training stage does code data help llms reasoning?
\newblock \emph{arXiv preprint arXiv:2309.16298}.

\bibitem[{Madaan et~al.(2024)Madaan, Tandon, Gupta, Hallinan, Gao, Wiegreffe, Alon, Dziri, Prabhumoye, Yang et~al.}]{madaan2023self}
Aman Madaan, Niket Tandon, Prakhar Gupta, Skyler Hallinan, Luyu Gao, Sarah Wiegreffe, Uri Alon, Nouha Dziri, Shrimai Prabhumoye, Yiming Yang, et~al. 2024.
\newblock Self-refine: Iterative refinement with self-feedback.
\newblock \emph{Advances in Neural Information Processing Systems}, 36.

\bibitem[{Mari et~al.(2013)Mari, Beyssade, and Del~Prete}]{mari2013genericity}
Alda Mari, Claire Beyssade, and Fabio Del~Prete. 2013.
\newblock \emph{Genericity}.
\newblock 43. Oxford University Press.

\bibitem[{McCarthy(1986)}]{mccarthy1986applications}
John McCarthy. 1986.
\newblock Applications of circumscription to formalizing common-sense knowledge.
\newblock \emph{Artificial intelligence}, 28(1):89--116.

\bibitem[{MistralAI(2023)}]{mixtral}
MistralAI. 2023.
\newblock \href {https://mistral.ai/news/mixtral-of-experts/} {Mixtral of experts - a high quality sparse mixture-of-experts}.

\bibitem[{NousResearch(2023)}]{openhermes}
NousResearch. 2023.
\newblock \href {https://huggingface.co/teknium/OpenHermes-2.5-Mistral-7B} {Openhermes 2.5 - mistral 7b}.

\bibitem[{Pan et~al.(2024)Pan, Saxon, Xu, Nathani, Wang, and Wang}]{pan2024automatically}
Liangming Pan, Michael Saxon, Wenda Xu, Deepak Nathani, Xinyi Wang, and William~Yang Wang. 2024.
\newblock Automatically correcting large language models: Surveying the landscape of diverse automated correction strategies.
\newblock \emph{Transactions of the Association for Computational Linguistics}, 12:484--506.

\bibitem[{Parmar et~al.(2024)Parmar, Patel, Varshney, Nakamura, Luo, Mashetty, Mitra, and Baral}]{parmar2024towards}
Mihir Parmar, Nisarg Patel, Neeraj Varshney, Mutsumi Nakamura, Man Luo, Santosh Mashetty, Arindam Mitra, and Chitta Baral. 2024.
\newblock Towards systematic evaluation of logical reasoning ability of large language models.
\newblock \emph{arXiv preprint arXiv:2404.15522}.

\bibitem[{Paul et~al.(2023)Paul, Ismayilzada, Peyrard, Borges, Bosselut, West, and Faltings}]{paul2023refiner}
Debjit Paul, Mete Ismayilzada, Maxime Peyrard, Beatriz Borges, Antoine Bosselut, Robert West, and Boi Faltings. 2023.
\newblock Refiner: Reasoning feedback on intermediate representations.
\newblock \emph{arXiv preprint arXiv:2304.01904}.

\bibitem[{Pelletier and Asher(1997)}]{pelletier1997generics}
Francis~Jeffry Pelletier and Nicholas Asher. 1997.
\newblock Generics and defaults.
\newblock In \emph{Handbook of logic and language}, pages 1125--1177. Elsevier.

\bibitem[{Perez et~al.(2023)Perez, Ringer, Lukosiute, Nguyen, Chen, Heiner, Pettit, Olsson, Kundu, Kadavath, Jones, Chen, Mann, Israel, Seethor, McKinnon, Olah, Yan, Amodei, Amodei, Drain, Li, Tran-Johnson, Khundadze, Kernion, Landis, Kerr, Mueller, Hyun, Landau, Ndousse, Goldberg, Lovitt, Lucas, Sellitto, Zhang, Kingsland, Elhage, Joseph, Mercado, DasSarma, Rausch, Larson, McCandlish, Johnston, Kravec, El~Showk, Lanham, Telleen-Lawton, Brown, Henighan, Hume, Bai, Hatfield-Dodds, Clark, Bowman, Askell, Grosse, Hernandez, Ganguli, Hubinger, Schiefer, and Kaplan}]{perez-etal-2023-discovering}
Ethan Perez, Sam Ringer, Kamile Lukosiute, Karina Nguyen, Edwin Chen, Scott Heiner, Craig Pettit, Catherine Olsson, Sandipan Kundu, Saurav Kadavath, Andy Jones, Anna Chen, Benjamin Mann, Brian Israel, Bryan Seethor, Cameron McKinnon, Christopher Olah, Da~Yan, Daniela Amodei, Dario Amodei, Dawn Drain, Dustin Li, Eli Tran-Johnson, Guro Khundadze, Jackson Kernion, James Landis, Jamie Kerr, Jared Mueller, Jeeyoon Hyun, Joshua Landau, Kamal Ndousse, Landon Goldberg, Liane Lovitt, Martin Lucas, Michael Sellitto, Miranda Zhang, Neerav Kingsland, Nelson Elhage, Nicholas Joseph, Noemi Mercado, Nova DasSarma, Oliver Rausch, Robin Larson, Sam McCandlish, Scott Johnston, Shauna Kravec, Sheer El~Showk, Tamera Lanham, Timothy Telleen-Lawton, Tom Brown, Tom Henighan, Tristan Hume, Yuntao Bai, Zac Hatfield-Dodds, Jack Clark, Samuel~R. Bowman, Amanda Askell, Roger Grosse, Danny Hernandez, Deep Ganguli, Evan Hubinger, Nicholas Schiefer, and Jared Kaplan. 2023.
\newblock \href {https://doi.org/10.18653/v1/2023.findings-acl.847} {Discovering language model behaviors with model-written evaluations}.
\newblock In \emph{Findings of the Association for Computational Linguistics: ACL 2023}, pages 13387--13434, Toronto, Canada. Association for Computational Linguistics.

\bibitem[{Pyatkin et~al.(2023)Pyatkin, Hwang, Srikumar, Lu, Jiang, Choi, and Bhagavatula}]{pyatkin-etal-2023-clarifydelphi}
Valentina Pyatkin, Jena~D. Hwang, Vivek Srikumar, Ximing Lu, Liwei Jiang, Yejin Choi, and Chandra Bhagavatula. 2023.
\newblock \href {https://doi.org/10.18653/v1/2023.acl-long.630} {{C}larify{D}elphi: Reinforced clarification questions with defeasibility rewards for social and moral situations}.
\newblock In \emph{Proceedings of the 61st Annual Meeting of the Association for Computational Linguistics (Volume 1: Long Papers)}, pages 11253--11271, Toronto, Canada. Association for Computational Linguistics.

\bibitem[{Ralethe and Buys(2022)}]{ralethe2022generic}
Sello Ralethe and Jan Buys. 2022.
\newblock Generic overgeneralization in pre-trained language models.
\newblock In \emph{Proceedings of the 29th International Conference on Computational Linguistics}, pages 3187--3196.

\bibitem[{Ranaldi and Pucci(2023)}]{ranaldi2023sycophancy}
Leonardo Ranaldi and Giulia Pucci. 2023.
\newblock \href {http://arxiv.org/abs/2311.09410} {When large language models contradict humans? large language models' sycophantic behaviour}.

\bibitem[{Rao et~al.(2023)Rao, Jiang, Pyatkin, Gu, Tandon, Dziri, Brahman, and Choi}]{rao2023makes}
Kavel Rao, Liwei Jiang, Valentina Pyatkin, Yuling Gu, Niket Tandon, Nouha Dziri, Faeze Brahman, and Yejin Choi. 2023.
\newblock \href {https://doi.org/10.18653/v1/2023.findings-emnlp.812} {What makes it ok to set a fire? iterative self-distillation of contexts and rationales for disambiguating defeasible social and moral situations}.
\newblock In \emph{Findings of the Association for Computational Linguistics: EMNLP 2023}, pages 12140--12159, Singapore. Association for Computational Linguistics.

\bibitem[{Reiter(1988)}]{reiter1988nonmonotonic}
Raymond Reiter. 1988.
\newblock Nonmonotonic reasoning.
\newblock In \emph{Exploring artificial intelligence}, pages 439--481. Elsevier.

\bibitem[{Ribeiro et~al.(2020)Ribeiro, Wu, Guestrin, and Singh}]{ribeiro-etal-2020-beyond}
Marco~Tulio Ribeiro, Tongshuang Wu, Carlos Guestrin, and Sameer Singh. 2020.
\newblock \href {https://doi.org/10.18653/v1/2020.acl-main.442} {Beyond accuracy: Behavioral testing of {NLP} models with {C}heck{L}ist}.
\newblock In \emph{Proceedings of the 58th Annual Meeting of the Association for Computational Linguistics}, pages 4902--4912, Online. Association for Computational Linguistics.

\bibitem[{R{\"o}ttger et~al.(2023)R{\"o}ttger, Kirk, Vidgen, Attanasio, Bianchi, and Hovy}]{rottger2023xstest}
Paul R{\"o}ttger, Hannah~Rose Kirk, Bertie Vidgen, Giuseppe Attanasio, Federico Bianchi, and Dirk Hovy. 2023.
\newblock Xstest: A test suite for identifying exaggerated safety behaviours in large language models.
\newblock \emph{arXiv preprint arXiv:2308.01263}.

\bibitem[{Rudinger et~al.(2020)Rudinger, Shwartz, Hwang, Bhagavatula, Forbes, Le~Bras, Smith, and Choi}]{rudinger2020thinking}
Rachel Rudinger, Vered Shwartz, Jena~D. Hwang, Chandra Bhagavatula, Maxwell Forbes, Ronan Le~Bras, Noah~A. Smith, and Yejin Choi. 2020.
\newblock \href {https://doi.org/10.18653/v1/2020.findings-emnlp.418} {Thinking like a skeptic: Defeasible inference in natural language}.
\newblock In \emph{Findings of the Association for Computational Linguistics: EMNLP 2020}, pages 4661--4675, Online. Association for Computational Linguistics.

\bibitem[{Russell(2001)}]{russell2001cognitive}
James Russell. 2001.
\newblock Cognitive theories of autism.
\newblock In \emph{Cognitive deficits in brain disorders}, pages 309--338. CRC Press.

\bibitem[{Shi et~al.(2023)Shi, Chen, Misra, Scales, Dohan, Chi, Sch{\"a}rli, and Zhou}]{shi2023large}
Freda Shi, Xinyun Chen, Kanishka Misra, Nathan Scales, David Dohan, Ed~H Chi, Nathanael Sch{\"a}rli, and Denny Zhou. 2023.
\newblock Large language models can be easily distracted by irrelevant context.
\newblock In \emph{International Conference on Machine Learning}, pages 31210--31227. PMLR.

\bibitem[{Shinn et~al.(2023)Shinn, Cassano, Gopinath, Narasimhan, and Yao}]{shinn2023reflexion}
Noah Shinn, Federico Cassano, Ashwin Gopinath, Karthik~R Narasimhan, and Shunyu Yao. 2023.
\newblock Reflexion: Language agents with verbal reinforcement learning.
\newblock In \emph{Thirty-seventh Conference on Neural Information Processing Systems}.

\bibitem[{Sloman and Lagnado(2005)}]{sloman2005problem}
Steven~A Sloman and David Lagnado. 2005.
\newblock The problem of induction.
\newblock \emph{The Cambridge handbook of thinking and reasoning}, pages 95--116.

\bibitem[{Stechly et~al.(2024)Stechly, Valmeekam, and Kambhampati}]{stechly2024self}
Kaya Stechly, Karthik Valmeekam, and Subbarao Kambhampati. 2024.
\newblock On the self-verification limitations of large language models on reasoning and planning tasks.
\newblock \emph{arXiv preprint arXiv:2402.08115}.

\bibitem[{Stenning and Van~Lambalgen(2012)}]{stenning2012human}
Keith Stenning and Michiel Van~Lambalgen. 2012.
\newblock \emph{Human reasoning and cognitive science}.
\newblock MIT Press.

\bibitem[{Touvron et~al.(2023)Touvron, Martin, Stone, Albert, Almahairi, Babaei, Bashlykov, Batra, Bhargava, Bhosale et~al.}]{touvron2023llama}
Hugo Touvron, Louis Martin, Kevin Stone, Peter Albert, Amjad Almahairi, Yasmine Babaei, Nikolay Bashlykov, Soumya Batra, Prajjwal Bhargava, Shruti Bhosale, et~al. 2023.
\newblock Llama 2: Open foundation and fine-tuned chat models.
\newblock \emph{arXiv preprint arXiv:2307.09288}.

\bibitem[{Tunstall et~al.(2023)Tunstall, Beeching, Lambert, Rajani, Rasul, Belkada, Huang, von Werra, Fourrier, Habib, Sarrazin, Sanseviero, Rush, and Wolf}]{tunstall2023zephyr}
Lewis Tunstall, Edward Beeching, Nathan Lambert, Nazneen Rajani, Kashif Rasul, Younes Belkada, Shengyi Huang, Leandro von Werra, Clémentine Fourrier, Nathan Habib, Nathan Sarrazin, Omar Sanseviero, Alexander~M. Rush, and Thomas Wolf. 2023.
\newblock \href {http://arxiv.org/abs/2310.16944} {Zephyr: Direct distillation of lm alignment}.

\bibitem[{Valmeekam et~al.(2023)Valmeekam, Marquez, Olmo, Sreedharan, and Kambhampati}]{valmeekam2023planbench}
Karthik Valmeekam, Matthew Marquez, Alberto Olmo, Sarath Sreedharan, and Subbarao Kambhampati. 2023.
\newblock Planbench: an extensible benchmark for evaluating large language models on planning and reasoning about change.
\newblock In \emph{Thirty-seventh Conference on Neural Information Processing Systems Datasets and Benchmarks Track}.

\bibitem[{Wang et~al.(2023{\natexlab{a}})Wang, Yue, and Sun}]{wang2023can}
Boshi Wang, Xiang Yue, and Huan Sun. 2023{\natexlab{a}}.
\newblock \href {https://doi.org/10.18653/v1/2023.findings-emnlp.795} {Can {C}hat{GPT} defend its belief in truth? evaluating {LLM} reasoning via debate}.
\newblock In \emph{Findings of the Association for Computational Linguistics: EMNLP 2023}, pages 11865--11881, Singapore. Association for Computational Linguistics.

\bibitem[{Wang et~al.(2023{\natexlab{b}})Wang, Xixu, Hou, Chen, Zheng, Wang, Yang, Ye, Huang, Geng et~al.}]{wang2023robustness}
Jindong Wang, HU~Xixu, Wenxin Hou, Hao Chen, Runkai Zheng, Yidong Wang, Linyi Yang, Wei Ye, Haojun Huang, Xiubo Geng, et~al. 2023{\natexlab{b}}.
\newblock On the robustness of chatgpt: An adversarial and out-of-distribution perspective.
\newblock In \emph{ICLR 2023 Workshop on Trustworthy and Reliable Large-Scale Machine Learning Models}.

\bibitem[{Wang et~al.(2022)Wang, Wei, Schuurmans, Le, Chi, Narang, Chowdhery, and Zhou}]{wang2022self}
Xuezhi Wang, Jason Wei, Dale Schuurmans, Quoc~V Le, Ed~H Chi, Sharan Narang, Aakanksha Chowdhery, and Denny Zhou. 2022.
\newblock Self-consistency improves chain of thought reasoning in language models.
\newblock In \emph{The Eleventh International Conference on Learning Representations}.

\bibitem[{Webson and Pavlick(2022)}]{webson-pavlick-2022-prompt}
Albert Webson and Ellie Pavlick. 2022.
\newblock \href {https://doi.org/10.18653/v1/2022.naacl-main.167} {Do prompt-based models really understand the meaning of their prompts?}
\newblock In \emph{Proceedings of the 2022 Conference of the North American Chapter of the Association for Computational Linguistics: Human Language Technologies}, pages 2300--2344, Seattle, United States. Association for Computational Linguistics.

\bibitem[{Wei et~al.(2022)Wei, Bosma, Zhao, Guu, Yu, Lester, Du, Dai, and Le}]{wei2022finetuned}
Jason Wei, Maarten Bosma, Vincent~Y. Zhao, Kelvin Guu, Adams~Wei Yu, Brian Lester, Nan Du, Andrew~M. Dai, and Quoc~V. Le. 2022.
\newblock \href {http://arxiv.org/abs/2109.01652} {Finetuned language models are zero-shot learners}.

\bibitem[{Wilcoxon(1992)}]{wilcoxon1992individual}
Frank Wilcoxon. 1992.
\newblock Individual comparisons by ranking methods.
\newblock In \emph{Breakthroughs in Statistics: Methodology and Distribution}, pages 196--202. Springer.

\bibitem[{Xie et~al.(2023)Xie, Zhang, Chen, Lou, and Su}]{xie2023adaptive}
Jian Xie, Kai Zhang, Jiangjie Chen, Renze Lou, and Yu~Su. 2023.
\newblock Adaptive chameleon or stubborn sloth: Revealing the behavior of large language models in knowledge conflicts.
\newblock In \emph{The Twelfth International Conference on Learning Representations}.

\bibitem[{Xu et~al.(2023)Xu, Sun, Zheng, Geng, Zhao, Feng, Tao, and Jiang}]{xu2023wizardlm}
Can Xu, Qingfeng Sun, Kai Zheng, Xiubo Geng, Pu~Zhao, Jiazhan Feng, Chongyang Tao, and Daxin Jiang. 2023.
\newblock \href {http://arxiv.org/abs/2304.12244} {Wizardlm: Empowering large language models to follow complex instructions}.

\bibitem[{Yang et~al.(2024)Yang, Liu, Wu, Yang, Fung, Li, Huang, Cao, Wang, Wang et~al.}]{yang2024if}
Ke~Yang, Jiateng Liu, John Wu, Chaoqi Yang, Yi~R Fung, Sha Li, Zixuan Huang, Xu~Cao, Xingyao Wang, Yiquan Wang, et~al. 2024.
\newblock If llm is the wizard, then code is the wand: A survey on how code empowers large language models to serve as intelligent agents.
\newblock \emph{arXiv preprint arXiv:2401.00812}.

\bibitem[{Zhu et~al.(2023{\natexlab{a}})Zhu, Frick, Wu, Zhu, and Jiao}]{starling2023}
Banghua Zhu, Evan Frick, Tianhao Wu, Hanlin Zhu, and Jiantao Jiao. 2023{\natexlab{a}}.
\newblock Starling-7b: Improving llm helpfulness \& harmlessness with rlaif.

\bibitem[{Zhu et~al.(2023{\natexlab{b}})Zhu, Wang, Zhou, Wang, Chen, Wang, Yang, Ye, Gong, Zhang et~al.}]{zhu2023promptbench}
Kaijie Zhu, Jindong Wang, Jiaheng Zhou, Zichen Wang, Hao Chen, Yidong Wang, Linyi Yang, Wei Ye, Neil~Zhenqiang Gong, Yue Zhang, et~al. 2023{\natexlab{b}}.
\newblock Promptbench: Towards evaluating the robustness of large language models on adversarial prompts.
\newblock \emph{arXiv preprint arXiv:2306.04528}.

\bibitem[{Ziems et~al.(2023)Ziems, Dwivedi-Yu, Wang, Halevy, and Yang}]{ziems-etal-2023-normbank}
Caleb Ziems, Jane Dwivedi-Yu, Yi-Chia Wang, Alon Halevy, and Diyi Yang. 2023.
\newblock \href {https://doi.org/10.18653/v1/2023.acl-long.429} {{N}orm{B}ank: A knowledge bank of situational social norms}.
\newblock In \emph{Proceedings of the 61st Annual Meeting of the Association for Computational Linguistics (Volume 1: Long Papers)}, pages 7756--7776, Toronto, Canada. Association for Computational Linguistics.

\end{thebibliography}

\appendix

\section{Example input}
\label{sec:example_input}

Example input formatted using our prompt template and the chat template\footnote{\url{https://huggingface.co/blog/chat-templates}}, e.g., for Mistral-7B-Instruct:\\

\noindent\fbox{\begin{minipage}{19em}
\underline{Prompt:}\\

\textit{[INST] Do you believe that the following statement is accurate: `Birds fly.'}\\

\textit{Please answer yes or no. [/INST]}\\
\textit{yes}\\
\textit{[INST] Penguins do not fly.}\\

\textit{Do you believe that the following statement is accurate: `Birds fly.'}\\

\textit{Please answer yes or no. [/INST]}
\end{minipage}}

\section{Additional information on data preprocessing}\label{sec:preprocessing}

For GEN-comm, we conduct additional processing to obtain high quality generics and ensure a parallel experimental setup between GEN-comm and GEN-abs. We retain only generics that were annotated as `valid' by human annotators. We filter generics for which both an exception and an instantiation exists. Since generics are unquantified statements, we remove any quantifiers such as `generally', `usually' and `typically' at the beginning of each generic. 
To enable consistent evaluation on GEN-abs and GEN-comm, we evaluate each LLM on generics contained in GEN-comm which it accepts \textit{a priori}. In an initial experiment, we prompt LLMs using the first part of our template (above; App.~\ref{sec:example_input}). An example input for \texttt{GEN-comm} would be, e.g., \textit{`[INST] Do you believe that the following statement is accurate: `Birds have property P.' Please answer yes or no[/INST]'}. Generics for which an LLM does not generate \textit{yes} as a response are discarded. We retain $>1200$ samples for each model~(See Table \ref{tab:num_generics} for details).

\begin{table}[t]
    \centering
    \begin{tabular}{l|c}
        Model & \# samples\\
        \hline
        Mistral-7B-Instruct & 2093 \\
        Llama-2-13b & 1245 \\
        Zephyr-7b-beta & 1536 \\
        WizardLM-13B-V1.2 & 2225 \\
        OpenHermes-2.5-Mistral-7B & 2153 \\
        Starling-LM-7B-alpha & 2244 \\
        Mixtral-8x7B-Instruct-v0.1 & 1959 \\
    \end{tabular}
    \caption{\# retained samples in \texttt{GEN-comm}}
    \label{tab:num_generics}
\end{table}

\begin{figure}[t]
    \centering
    \includegraphics[width=.47\textwidth]{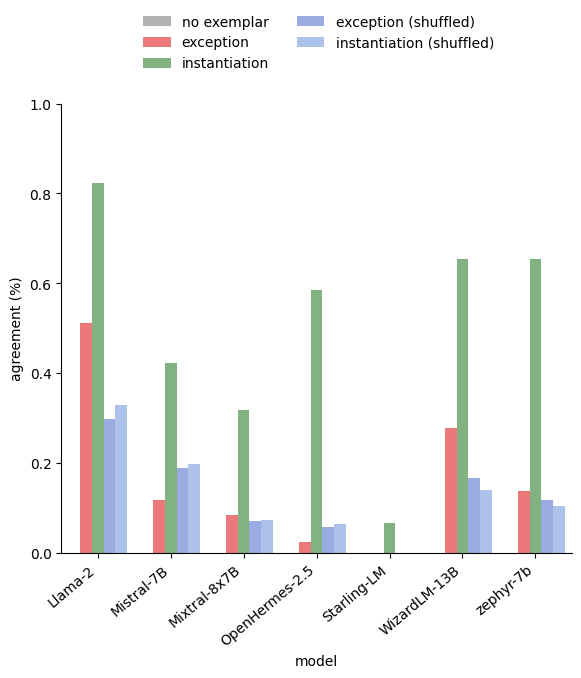}
    \caption{Results on generics contained in \texttt{GEN-comm} that are rejected a priori. Missing bars for `no exemplar' indicate agreement rates of zero.}
    \label{fig:results_penguins_dialogue_reject}
\end{figure}

Results on the resultant dataset are presented in the main body of the paper (Section \ref{results}). For the reader's interest, we include here  also LLM responses to generics contained in \texttt{GEN-comm} which are rejected by LLMs, i.e., a given LLM generates the response \textit{no} to the prompt above (See Figure \ref{fig:results_penguins_dialogue_reject}). As expected agreement rates soar for almost all models when adding an instantiation which confirms the previously rejected generic. Nevertheless, agreement rates also increase, albeit less, when adding \textit{exceptions} or unrelated random examplars, particularly for Llama-2 and WizardLM. OpenHermes and Starling show the least inconsistencies.

\section{Additional information on LLMs}\label{sec:appendix_models}

In this section we provide additional details on the models used in this study which are listed in Section \ref{models}.
The specific checkpoints we use can be seen in Table \ref{tab:checkpoints} and are all available through the HuggingFace Hub. All models we use are trained for chat interaction. 
\begin{table}[t]
    \centering
    \begin{tabular}{l}
        LLM Checkpoints\\
        \hline
        meta-llama/Llama-2-13b-chat-hf\\
        mistralai/Mistral-7B-Instruct-v0.2\\
        mistralai/Mixtral-8x7B-Instruct-v0.1\\
        HuggingFaceH4/zephyr-7b-beta\\
        berkeley-nest/Starling-LM-7B-alpha\\
        WizardLM/WizardLM-13B-V1.2\\
        teknium/OpenHermes-2.5-Mistral-7B
    \end{tabular}
    \caption{LLM checkpoints used in this study.}
    \label{tab:checkpoints}
\end{table}

Mixtral-8x7B-Instruct-v0.1 \citep{mixtral} is a sparse mixture of expert model based on 8 Mistral 7B models that has been further trained using supervised finetuning and Direct Preference Optimisation. It ranks highest among its weight class on AlpacaEval\footnote{\url{https://tatsu-lab.github.io/alpaca_eval/}} and chat.lmsys\footnote{\url{https://chat.lmsys.org/?leaderboard}} leaderboards (as of Feb 6 2024). At its release it surpasses GPT-3.5 and LLaMA-2-70b.

StarlingLM-13B-V1.2 \citep{starling2023} has been trained via Reinforcement Learning from AI Feedback (RLAIF) on the Nectar dataset. In its weight class, it is the second best performing model on chat.lmsys and 4th on AlpacaEval (as of Feb 6 2024).

Amidst mounting evidence that training on code enhances reasoning abilities also for natural language \citep{liang2022helm,yang2024if,ma2023training}, we also use OpenHermes-2.5-Mistral-7B \cite{openhermes} which ranks third in its weight class on chat.lmsys. It is Mistral-based model that has been finetuned on additional code datasets. Notably, the developers detail that this results in improvements on non-code tasks.\footnote{\url{https://huggingface.co/teknium/OpenHermes-2.5-Mistral-7B}}

WizardLM-13B-V1.2 \citep{xu2023wizardlm} is a finetuned version of Llama-2 13b and is ranked 8th in its weight-class on both chat.lmsys and AlpacaEval.

Zephyr-7b-beta \citep{tunstall2023zephyr} is a finetuned version of Mistral-7B-v0.1. It is ranked 9th on chat.lmsys and 11th on AlpacaEval. 

\section{Average runtime}
Generating LLM responses for one LLM and all generics across all settings took less than $0.5$ GPU hours. All experiments were conducted on one NVIDIA A100 GPU.

\section{Statistical test results}\label{app:statisticaltestresults}

Responses in the presence of exemplars are significantly different from results obtained without examplars (see Tables \ref{tab:wilcoxon}, \ref{tab:wilcoxon_with}, \ref{tab:wilcoxon_cot}), for all types of exemplars and all models (significance level $0.01$; sole exception is Llama-2 with CoT prompting as can be seen in Table \ref{tab:wilcoxon_cot} rows 1-2).

\begin{figure}[t]
    \centering
    \includegraphics[width=.47\textwidth]{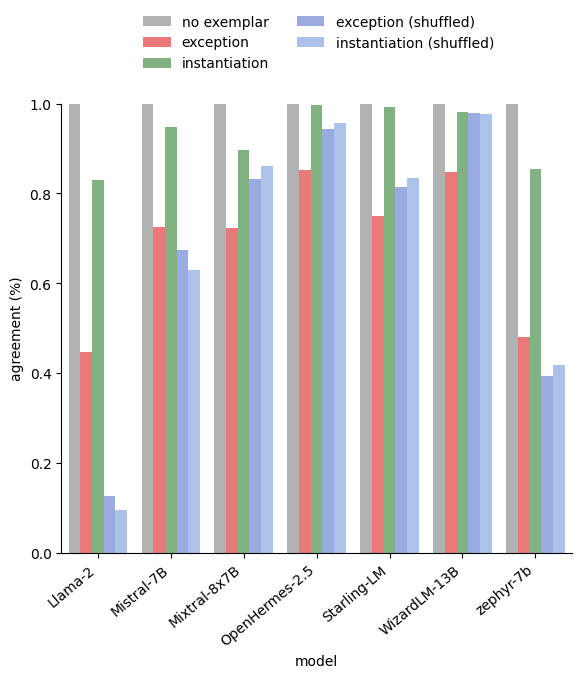}
    \caption{Results on \texttt{GEN-comm}. Alternative prompt template described in Section \ref{appendix_additional_results}}
    \label{fig:results_penguins_dialogue}
\end{figure}

\begin{figure}[t]
    \centering
    \includegraphics[width=.47\textwidth]{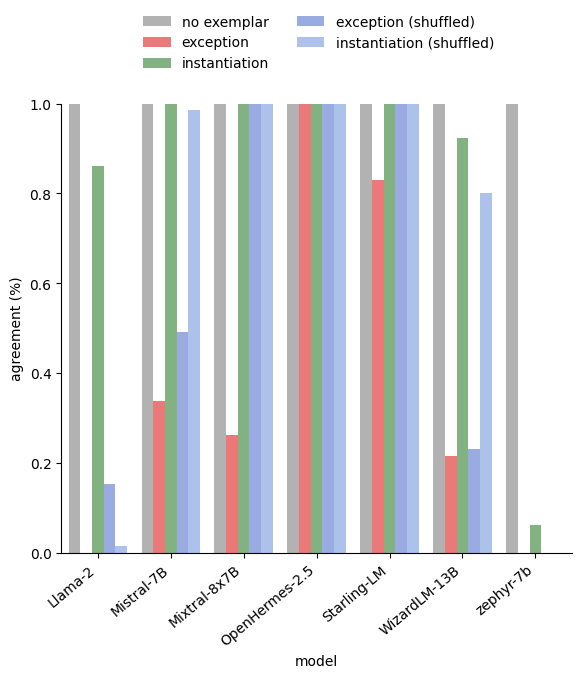}
    \caption{Results on \texttt{GEN-abs}. Alternative prompt template described in Section \ref{appendix_additional_results}.}
    \label{fig:results_han_dialogue}
\end{figure}

\begin{figure}[t]
    \centering
    \includegraphics[width=.47\textwidth]{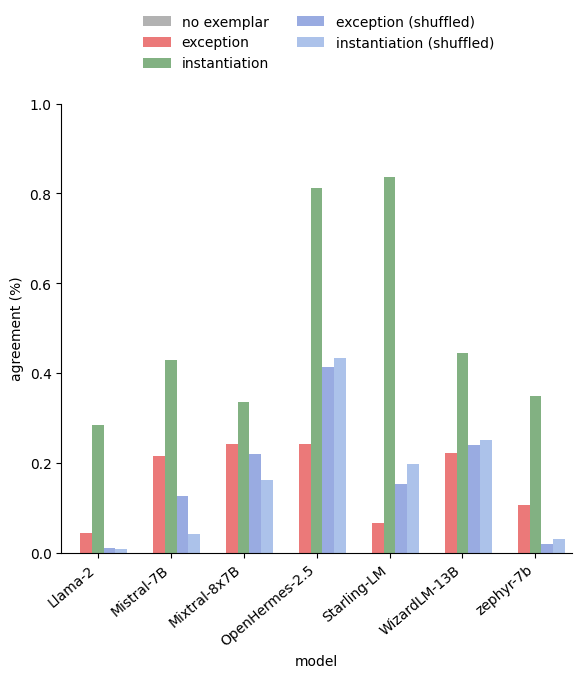}
    \caption{Results on generics of \texttt{GEN-comm} that are rejected by LLMs a priori. Alternative prompt template described in Section \ref{appendix_additional_results}. Missing bars indicate that agreement for `no exemplar' is zero.}
    \label{fig:results_han_dialogue_reject}
\end{figure}

\section{Additional experimental results}\label{appendix_additional_results}

We demonstrate additional experimental results based on an alternative prompting set-up in Figures \ref{fig:results_penguins_dialogue} and \ref{fig:results_han_dialogue}.

To this end, we prompt LLMs using the following template where [INST] is an example of a model-specific special token used in chat templating. For example: \\

\noindent\fbox{\begin{minipage}{19em}
\underline{Prompt}\\
\textit{[INST] Do you believe that the following statement is accurate: `Birds fly'}\\

\textit{Please answer yes or no. [/INST]}
\end{minipage}}\\

For \texttt{GEN-comm}, we retain all generics to which an LLM responds \textit{yes} to the prompt above. We then prompt LLMs anew supplying an exception, instantiation or random exemplar together with a generic for both datasets. For example:\\

\noindent\fbox{\begin{minipage}{19em}
\underline{Prompt}\\
\textit{[INST] Penguins do not fly.}\\

\textit{Do you believe that the following statement is accurate: `Birds fly'}\\

\textit{Please answer yes or no. [/INST]}
\end{minipage}}\\

We find that results differ significantly between the two conditions (no exemplar vs. with an exemplar) (see Table \ref{tab:wilcoxon_with} for statistical test results). On \texttt{GEN-comm} (Figure \ref{fig:results_penguins_dialogue}) agreement rates drop considerably in the presence of exceptions which mirrors nonmonotonic reasoning patterns. Agreement is higher, yet still drops significantly in the presence of instantiations. No LLM maintains perfectly consistent responses at the addition of random instantiations or exceptions. When prompting with random exemplars surprisingly agreement drops, most notably for Llama-2 and Zephyr. 

For the reader's interest, we also include results on the portion of generics in \texttt{GEN-comm} which is rejected by LLMs a priori (Table \ref{fig:results_han_dialogue_reject}). As expected, agreement increases from zero at the addition of an instantiation to the prompt, most notably for OpenHermes and Starling. However, LLMs should maintain a response of \textit{no} at the addition of an \textit{exception} or random exemplar to the prompt. This is visibly not the case with agreement rates increasing significantly for all models. 

On \texttt{GEN-abs}, agreement drops considerably at the addition of an exception for all models except OpenHermes (Figure \ref{fig:results_han_dialogue}). Notably OpenHermes and Starling-LM appear to yield consistent responses in the presence of our controls, the random exemplars, while Llama-2 and Zephyr perform worst in that regard.

\subsection{Chain-of-thought prompting}

\begin{figure}
    \centering
    \includegraphics[width=.47\textwidth]{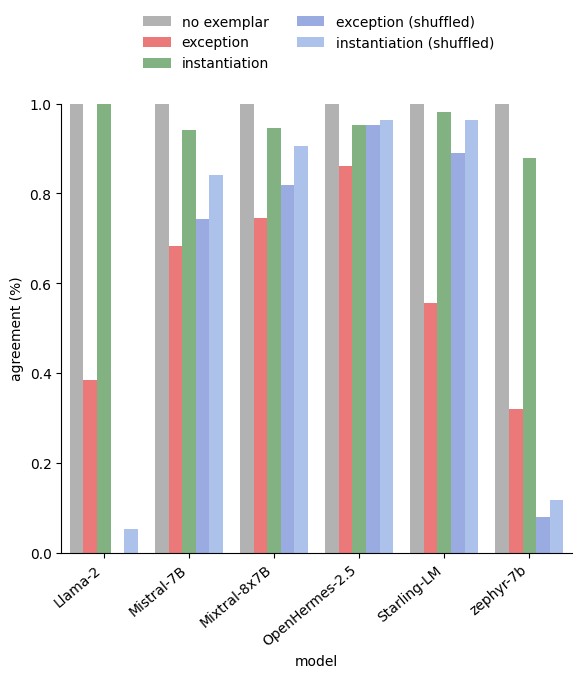}
    \caption{Results on \texttt{GEN-comm} using zero-shot CoT prompting.}
    \label{fig:cot-penguins}
\end{figure}

\begin{figure}
    \centering
    \includegraphics[width=.47\textwidth]{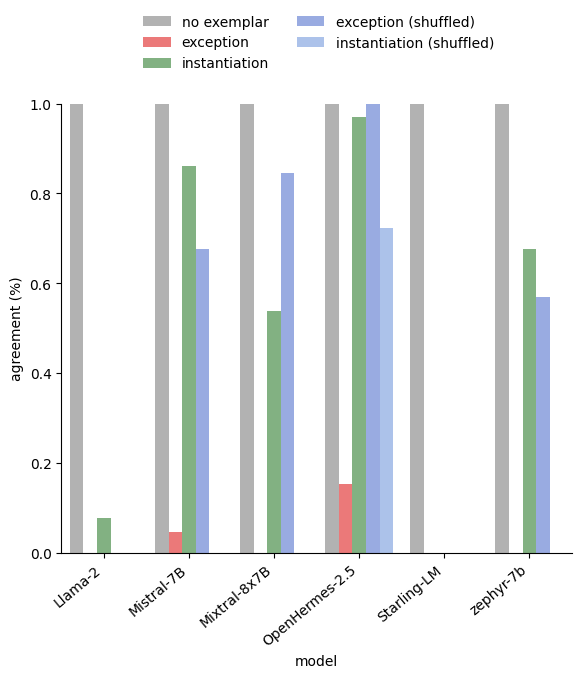}
    \caption{Results on \texttt{GEN-abs} using zero-shot CoT prompting. Missing bars indicate agreement rate of $0\%$.}
    \label{fig:cot-han}
\end{figure}

Additionally, we ran experiments using zero-shot Chain-of-Thought (CoT) prompting in the style of \citep{kojima2022large} by appending `Let's think step by step' to our prompts. We present results on \texttt{GEN-comm} in Figure \ref{fig:cot-penguins} and results on \texttt{GEN-abs} in Figure \ref{fig:cot-han}.

On \texttt{GEN-comm}, agreement rates drop significantly for all models at the addition of exceptions, instantiations or shuffled exemplars (with the exception of Llama-2 when we include instantiations;  see Table \ref{tab:wilcoxon_cot} for significance results). Agreement rates drop more given exceptions in comparison to instantiations or unrelated examplars for Mistral, Mixtral, OpenHermes and Starling. For Llama-2 and Zephyr agreement rates fall below $10\%$ at the addition of unrelated exemplars. 

On \texttt{GEN-abs}, agreement rates fall drastically given exceptions and equal $0\%$ for Llama-2, Mixtral, Starling and Zephyr. The same is true for shuffled instantiations. OpenHermes is the only model to maintain agreement rates above $90\%$ when presented with instantiations or shuffled exceptions.

\begin{table*}[b]
    \centering
    \begin{tabular}{l|l|l}
    Model & prompt setting & p-value\\
    \hline
        Llama-2-13b-chat-hf & exception & 1.2444035588550786e-84\\
Llama-2-13b-chat-hf & instantiation & 1.3944889010907487e-28\\
Llama-2-13b-chat-hf & exception (shuffled) & 1.3664041679452567e-86\\
Llama-2-13b-chat-hf & instantiation (shuffled) & 3.7504271121760947e-128\\
OpenHermes-2.5-Mistral-7B & exception & 2.0884875837625446e-45\\
OpenHermes-2.5-Mistral-7B & instantiation & 7.237829871739995e-08\\
OpenHermes-2.5-Mistral-7B & exception (shuffled) & 1.733880104231141e-27\\
OpenHermes-2.5-Mistral-7B & instantiation (shuffled) & 9.799073841979368e-26\\
Starling-LM-7B-alpha & exception & 1.0691632340127197e-102\\
Starling-LM-7B-alpha & instantiation & 7.247101964362887e-14\\
Starling-LM-7B-alpha & exception (shuffled) & 3.14927364689666e-77\\
Starling-LM-7B-alpha & instantiation (shuffled) & 5.588400099286033e-62\\
Mixtral-8x7B-Instruct-v0.1 & exception & 5.599059901868063e-84\\
Mixtral-8x7B-Instruct-v0.1 & instantiation & 4.84145282763492e-53\\
Mixtral-8x7B-Instruct-v0.1 & exception (shuffled) & 1.8855259265259482e-119\\
Mixtral-8x7B-Instruct-v0.1 & instantiation (shuffled) & 3.312378211336223e-151\\
WizardLM-13B-V1.2 & exception & 3.169934685227252e-109\\
WizardLM-13B-V1.2 & instantiation & 1.244192114854348e-15\\
WizardLM-13B-V1.2 & exception (shuffled) & 6.7440576522393956e-49\\
WizardLM-13B-V1.2 & instantiation (shuffled) & 3.312389179997469e-50\\
zephyr-7b-beta & exception & 3.2434215158679907e-99\\
zephyr-7b-beta & instantiation & 2.68778179464934e-25\\
zephyr-7b-beta & exception (shuffled) & 2.7464111838608292e-137\\
zephyr-7b-beta & instantiation (shuffled) & 2.671546422248841e-187\\
Mistral-7B-Instruct-v0.2 & exception & 6.521923113646968e-71\\
Mistral-7B-Instruct-v0.2 & instantiation & 2.0670658180782593e-15\\
Mistral-7B-Instruct-v0.2 & exception (shuffled) & 6.923699393684986e-120\\
Mistral-7B-Instruct-v0.2 & instantiation (shuffled) & 4.9982887921763924e-139\\
    \end{tabular}
    \caption{Results of Wilcoxon signed ranked test for paired samples. We compare agreement of LLMs to generics with and without an exemplar (one of exception, instantiation, exception (shuffled), instantiation (shuffled). Results are obtained using the original prompt template described in section \ref{results} and correspond to the main results in the paper in Figure \ref{fig:results}.}
    \label{tab:wilcoxon}
\end{table*}

\begin{table*}
    \centering
    \begin{tabular}{l|l|l}
    Model & prompt setting & p-value\\
    \hline
    Llama-2-13b-chat-hf & exception & 1.2402659787920488e-62\\
Llama-2-13b-chat-hf & instantiation & 1.8577351435735865e-29\\
Llama-2-13b-chat-hf & exception (shuffled) & 6.558556037957885e-98\\
Llama-2-13b-chat-hf & instantiation (shuffled) & 9.990918651724453e-148\\
OpenHermes-2.5-Mistral-7B & exception & 9.041178413936276e-31\\
OpenHermes-2.5-Mistral-7B & instantiation & 0.025347318677468252\\
OpenHermes-2.5-Mistral-7B & exception (shuffled) & 9.236596617174027e-13\\
OpenHermes-2.5-Mistral-7B & instantiation (shuffled) & 1.2052982584446398e-13\\
Starling-LM-7B-alpha & exception & 4.84145282763492e-53\\
Starling-LM-7B-alpha & instantiation & 0.0009111188771537128\\
Starling-LM-7B-alpha & exception (shuffled) & 9.89884333064868e-40\\
Starling-LM-7B-alpha & instantiation (shuffled) & 6.7440576522393956e-49\\
Mixtral-8x7B-Instruct-v0.1 & exception & 2.6891242658680216e-51\\
Mixtral-8x7B-Instruct-v0.1 & instantiation & 2.8706760140807313e-27\\
Mixtral-8x7B-Instruct-v0.1 & exception (shuffled) & 7.287679729162835e-32\\
Mixtral-8x7B-Instruct-v0.1 & instantiation (shuffled) & 1.8712872006902566e-36\\
WizardLM-13B-V1.2 & exception & 5.8780179991539864e-33\\
WizardLM-13B-V1.2 & instantiation & 9.633570086430965e-07\\
WizardLM-13B-V1.2 & exception (shuffled) & 7.74421643104407e-06\\
WizardLM-13B-V1.2 & instantiation (shuffled) & 2.5802843041604163e-08\\
zephyr-7b-beta & exception & 3.525239394844374e-74\\
zephyr-7b-beta & instantiation & 2.476062658812572e-30\\
zephyr-7b-beta & exception (shuffled) & 3.7238080067294776e-86\\
zephyr-7b-beta & instantiation (shuffled) & 9.415767818703249e-116\\
Mistral-7B-Instruct-v0.2 & exception & 3.9328331793483447e-54\\
Mistral-7B-Instruct-v0.2 & instantiation & 2.0670658180782593e-15\\
Mistral-7B-Instruct-v0.2 & exception (shuffled) & 3.699479889932592e-64\\
Mistral-7B-Instruct-v0.2 & instantiation (shuffled) & 2.6476609044572044e-100\\
    \end{tabular}
    \caption{Results of Wilcoxon signed ranked test for paired samples. We compare agreement of LLMs to generics with and without an exemplar (one of exception, instantiation, exception (shuffled), instantiation (shuffled)). These results correspond to the alternative prompting style and results described in section \ref{appendix_additional_results}.}
    \label{tab:wilcoxon_with}
\end{table*}

\begin{table*}
    \centering
    \begin{tabular}{l|l|l}
    Model & prompt setting & p-value\\
    \hline
Llama-2-13b-chat-hf & exception & 0.025347318677468252\\
Llama-2-13b-chat-hf & instantiation & 0.31731050786291415\\
Llama-2-13b-chat-hf & exception (shuffled) & 0.0009111188771537128\\
Llama-2-13b-chat-hf & instantiation (shuffled) & 3.737981840170154e-05\\
Starling-LM-7B-alpha & exception & 4.320463057827488e-08\\
Starling-LM-7B-alpha & instantiation & 5.733031437583866e-07\\
Starling-LM-7B-alpha & exception (shuffled) & 1.5417257900279904e-08\\
Starling-LM-7B-alpha & instantiation (shuffled) & 1.1825298845719069e-11\\
OpenHermes-2.5-Mistral-7B & exception & 2.3159484001346495e-35\\
OpenHermes-2.5-Mistral-7B & instantiation & 3.552964224155306e-33\\
OpenHermes-2.5-Mistral-7B & exception (shuffled) & 4.4044942248007814e-32\\
OpenHermes-2.5-Mistral-7B & instantiation (shuffled) & 1.773177466197228e-41\\
Mixtral-8x7B-Instruct-v0.1 & exception & 2.9303133449994263e-53\\
Mixtral-8x7B-Instruct-v0.1 & instantiation & 4.474661339129513e-39\\
Mixtral-8x7B-Instruct-v0.1 & exception (shuffled) & 6.758775639492622e-37\\
Mixtral-8x7B-Instruct-v0.1 & instantiation (shuffled) & 5.058648827940248e-40\\
zephyr-7b-beta & exception & 3.6136286243610392e-96\\
zephyr-7b-beta & instantiation & 8.956226067732092e-94\\
zephyr-7b-beta & exception (shuffled) & 1.2813208444193637e-111\\
zephyr-7b-beta & instantiation (shuffled) & 2.0076004412348868e-151\\
Mistral-7B-Instruct-v0.2 & exception & 3.294362383314041e-67\\
Mistral-7B-Instruct-v0.2 & instantiation & 6.210993425425191e-19\\
Mistral-7B-Instruct-v0.2 & exception (shuffled) & 2.380470154600155e-54\\
Mistral-7B-Instruct-v0.2 & instantiation (shuffled) & 1.2444035588550786e-84\\
    \end{tabular}
    \caption{Results of Wilcoxon signed ranked test for paired samples. We compare agreement of LLMs to generics with and without an exemplar (one of exception, instantiation, exception (shuffled), instantiation (shuffled)). These results correspond to Chain-of-Thought prompting results described in section \ref{appendix_additional_results}.}
    \label{tab:wilcoxon_cot}
\end{table*}

\end{document}